\documentclass[journal]{IEEEtran}

\usepackage{graphicx}
\usepackage{subfigure}

\usepackage{cite}

\usepackage{amsmath,amsfonts}
\usepackage{algorithmic}
\usepackage{algorithm}
\usepackage{array}
\usepackage{textcomp}
\usepackage{stfloats}
\usepackage{url}
\usepackage{verbatim}
\usepackage{amssymb}
\usepackage{amsmath}
\usepackage{latexsym}
\usepackage{array}
\usepackage{url}
\usepackage{hyperref}
\usepackage{color} 
\usepackage{float}
\usepackage{epstopdf}
\usepackage{booktabs}
\usepackage{multirow}
\usepackage{lineno}

\begin{document}

\title{The Worse The Better: Content-Aware Viewpoint Generation Network for Projection-related Point Cloud Quality Assessment}

\author{Zhiyong~Su,
    Bingxu~Xie,
	Zheng~Li,  
	Jincan~Wu,
	and~Weiqing~Li
	\thanks{Zhiyong Su,  Bingxu Xie, Zheng Li,  and Jincan Wu are with the School of Automation, Nanjing University of Science and Technology, Nanjing, Jiangsu Province 210094, P.R. China (e-mail: su@njust.edu.cn, 2580801219@qq.com, 1315571938@qq.com,  2500381260@qq.com)} 
	
	\thanks{Weiqing Li is with the School of Computer Science and Engineering, Nanjing University of Science and Technology, Nanjing, Jiangsu Province 210094, P.R. China (e-mail: li\_weiqing@njust.edu.cn).}
	
	\thanks{Manuscript received 00 00, 0000; revised 00 00, 0000. This work was supported in part by National Key R\&D Program of China under Grant  2022QY0102, and Pre-research Project of The 14th Five Year Plan under Grant 50904040201 and Grant 2020-JCJQ-ZD-007-00. (Corresponding author: Weiqing Li.)}}

\markboth{Journal of \LaTeX\ Class Files,~Vol.~14, No.~8, August~2021}%
{Shell \MakeLowercase{\textit{et al.}}: A Sample Article Using IEEEtran.cls for IEEE Journals}

\IEEEpubid{\begin{minipage}{\textwidth}\ \centering
		Copyright \copyright 2025 IEEE. Personal use of this material is permitted. \\
		However, permission to use this material for any other purposes must be obtained 
		from the IEEE by sending an email to pubs-permissions@ieee.org.
\end{minipage}}
\maketitle

\begin{abstract}

Existing projection-related point cloud quality assessment (PCQA) methods commonly adopt a straightforward but content-independent projection strategy, which selects a certain number of viewpoints to obtain projected images of degraded point clouds for further assessment.
Through experimental studies, however, we observed the instability of final predicted quality scores, which change significantly over different viewpoint settings. 
Inspired by the "wooden barrel theory", given the default content-independent viewpoints of existing projection-related PCQA approaches, this paper presents a novel content-aware viewpoint generation network (CAVGN) to learn better viewpoints by taking the distribution of geometric and attribute features of degraded point clouds into consideration.
Firstly, the proposed CAVGN extracts multi-scale geometric and texture features of the entire input point cloud, respectively.
Then, for each default content-independent viewpoint, the extracted geometric and texture features are refined to focus on its corresponding visible part of the input point cloud.
Finally, the refined geometric and texture features are concatenated to generate an optimized viewpoint.
To train the proposed CAVGN, we present a self-supervised viewpoint ranking network (SSVRN) to select the viewpoint with the worst quality projected image to construct a default-optimized viewpoint dataset, which consists of thousands of paired default viewpoints and corresponding optimized viewpoints.
Experimental results show that the projection-related PCQA methods can achieve higher performance using the viewpoints generated by the proposed CAVGN.
The source code can be found at \href{https://github.com/yokeno1/CAVGN1}{https://github.com/yokeno1/CAVGN1}.
\end{abstract}


\section{Introduction}

\IEEEPARstart{W}{ith} the rapid development of geometric sensing technology, point clouds have been widely used in the fields of 3D modeling \cite{intro_2021cvpr_3dModeling,intro_2021cvpr_3dModeling2}, object recognition \cite{intro_2021cvpr_recognition1,intro_2021cvpr_recognition2}, navigation \cite{intro_2021cvpr_navigation,intro_2021cvpr_navigation2}, etc\cite{tian2020adaptive,xie2020point,wang2021multiple,que2021voxelcontext}.
Each point cloud consists of a set of unordered points that are described by geometric coordinates and optional attributes (e.g., RGB colors) \cite{gu20193d,yuan2020sampling}.
Due to the limitations of sensors or the content complexity of the actual scene, the collected raw point clouds usually contain noises. 
In addition, various distortions may be introduced during the processes of compression, transmission, and storage, leading to degraded geometric and perceptual quality \cite{su2019perceptual,Chai24lfi,Chai2022Monocular}.
However, most applications require high-quality point clouds that faithfully reflect the geometry and perceptual attributes of the physical world.
Therefore, objective point cloud quality assessment (PCQA) plays a crucial role in measuring the visual and geometric quality of distorted point clouds, and thus evaluating the performance of the corresponding point cloud related applications.

Generally, for the objective PCQA, there exist three categories: point-based methods \cite{Alexiou2020pointxr,viola2020color-based,meynet220FR-PCQA,Diniz2021CG,Zhou2024sgr}, projection-based methods \cite{yang82020-3d-2d,liu12-2021-PQA-net,yang13-2022-NR-domationadapt,Zhou23rr,zhang2023gms,xie2023pmBQA,Chen2024gcn}, and multimodal-based methods \cite{Chen2024DHC,zhang2023mmpcqa,liu2024mft}.
Both projection-based and multimodal-based PCQA methods are projection-related approaches.
Point-based PCQA approaches directly utilize geometric and attribute information to evaluate the quality of point clouds in 3D space \cite{zhang2022no}. 
Projection-based PCQA methods, which dominate the field of objective PCQA, project the 3D point cloud onto a preset number of projection regions to obtain corresponding 2D images through orthogonal \cite{liu12-2021-PQA-net} or perspective projection \cite{yang82020-3d-2d}, and then take full advantage of existing image quality assessment (IQA) metrics to assess the quality of point clouds in the image space \cite{ou2021sdd}.
Multimodal-based PCQA methods usually have two branches: one branch extracts 3D point features, the other branch extracts 2D projection features. 
The features extracted from the two branches are fused for the quality prediction.

\begin{figure}[!t]
	\centering
	\includegraphics[width=0.48\textwidth]{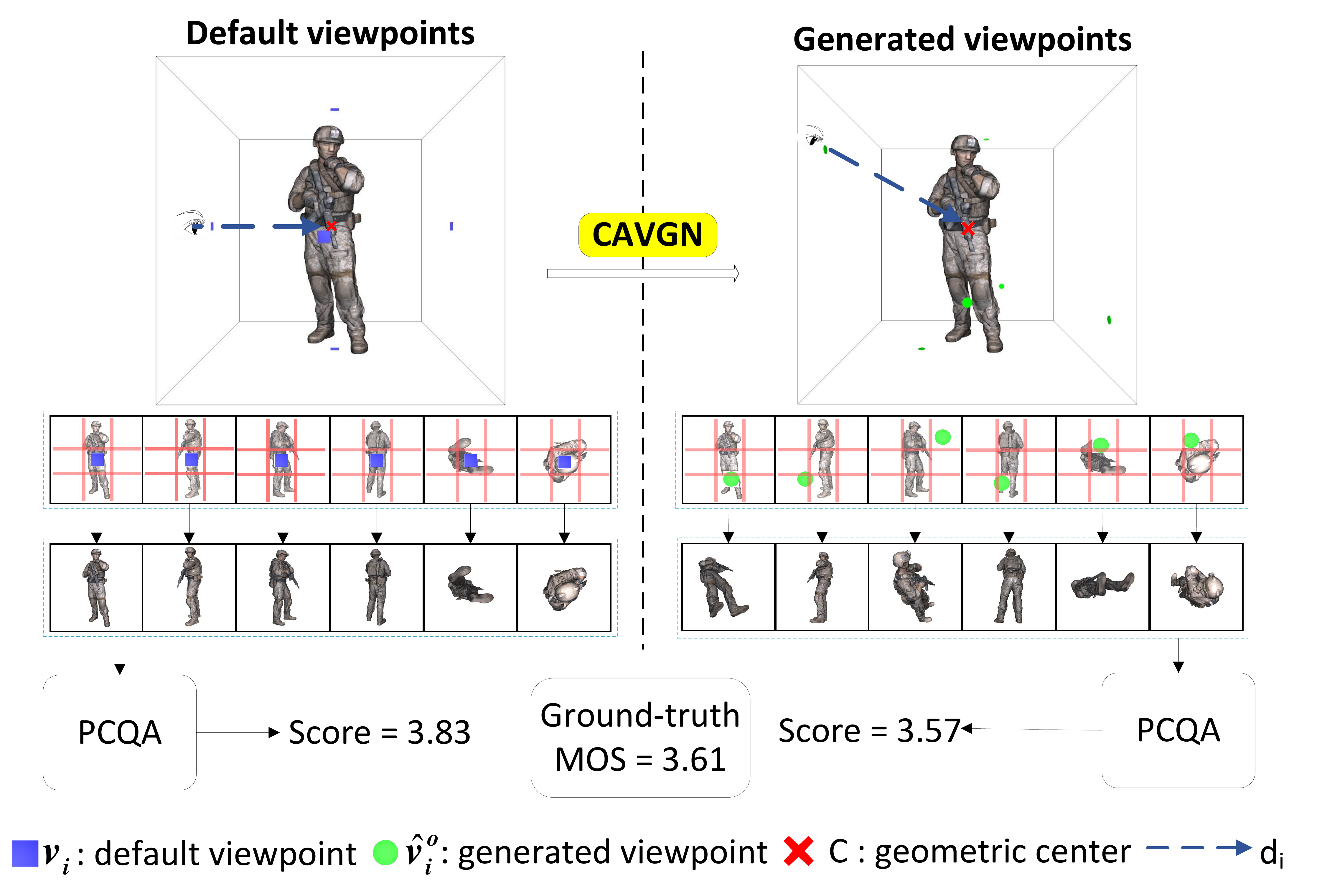}
	\caption{Comparison of default viewpoints and optimized viewpoints generated by CAVGN.}
	\label{fig:examples of shortcomings in traditional viewpoints}
\end{figure}

However, existing projection-related PCQA methods adopt a content-independent viewpoint setting strategy which does not take the distribution of geometric and attribute information of degraded point clouds into consideration.
Formally, given a degraded point cloud $\mathcal{P}$ with the geometric center $c$, the projection process of existing algorithms can be summarized as follows \cite{yang13-2022-NR-domationadapt,chen9-2021-layer-proj,yang82020-3d-2d,liu12-2021-PQA-net}.
Firstly, a specified number of viewpoints are initially set up, e.g., the commonly used cube-like viewpoint setting \cite{yang82020-3d-2d,yang13-2022-NR-domationadapt,liu12-2021-PQA-net}, as shown in Fig. \ref{fig:examples of shortcomings in traditional viewpoints}.
Then, for each viewpoint $v_i$, the projection direction $\mathbf{d}_i$ is defined as $\mathbf{d}_i = \overrightarrow{v_i - c}$.
After that, the projection region $r_i$ is defined as the region that passes through the viewpoint $v_i$ and is perpendicular to the projection direction $\mathbf{d}_i$ \cite{MPEG-VPCC,liu12-2021-PQA-net,yang82020-3d-2d}.
Note that $v_i$ is by default the center of $r_i$.
\IEEEpubidadjcol
Finally, the degraded point cloud $\mathcal{P}$ is mapped to each projection region $r_i$ according to the corresponding projection direction $\mathbf{d}_i$ through orthogonal projection \cite{liu12-2021-PQA-net} or perspective projection \cite{yang82020-3d-2d}.
In brief, each projection region $r_i$ corresponds to a visible part of $\mathcal{P}$, named $\mathcal{P}^{v_i}$ ($\mathcal{P}^{v_i} \in \mathcal{P}$) which is thus projected onto the image space according to the default viewpoint $v_i$ (i.e. projection direction $\mathbf{d}_i$).

Despite the progress made by existing projection-related PCQA methods, a natural question arises: given the projection region $r_i$ and its corresponding visible part $\mathcal{P}^{v_i}$, is there any viewpoint on $r_i$ which can better reflect the authentic quality of $\mathcal{P}^{v_i}$ than the default viewpoint $v_i$ ?
In other words, from which viewpoint is the observation a more complete reflection of the quality of $\mathcal{P}^{v_i}$ ?
Notably, we found out that existing content-independent viewpoint setting strategies have failed to consider the impact of the setting of viewpoints that intuitively affects the visual perception quality, leading to inferior assessment results \cite{liuyp23,Wang24eemf}.
They generally choose to set viewpoints (i.e. projection regions) by default, without considering the distribution of geometric and attribute information of degraded point clouds.
Consequently, it is difficult to obtain qualified projected images that can faithfully reflect the authentic quality of each $\mathcal{P}^{v_i}$ as well as the entire degraded point cloud $\mathcal{P}$.
To justify the observation, we draw inspiration from the "wooden barrel theory" which means that the capacity of a barrel is determined not by the longest wooden bars but by the shortest, and evaluate the impact of different viewpoint settings on the final prediction results.
As illustrated in Fig. \ref{fig:examples of shortcomings in traditional viewpoints}, as the setting of viewpoints changes, the quality score predicted by existing projection-related PCQA methods is also different. 
Unlike the default viewpoints used by existing methods, for each visible part $\mathcal{P}^{v_i}$ (i.e. projection region $r_i$), the viewpoint of the projected image with the lowest perceptual quality is selected as the generated viewpoint in this paper.
It can be observed that the PCQA methods with the generated viewpoints outperform the ones with default viewpoints, which just confirms the "wooden barrel theory".

Based on the above observation, this paper proposes a content-aware viewpoint generation network (CAVGN) to learn better viewpoints for existing projection-related PCQA methods, taking the distribution of geometric and attribute information into consideration.
Given a degraded point cloud $\mathcal{P}$, let the set of default viewpoints of existing projection-related PCQA methods be $\{ v_i \}$, the proposed CAVGN first extracts multi-scale geometric feature $F^g$ and texture feature $F^t$ of $\mathcal{P}$, respectively.
Then, for each default viewpoint $v_i$, the CAVGN refines $F^g$ and $F^t$ to focus on the partial point cloud $\mathcal{P}^{v_i}$ through encoding $v_i$ into $F^g$ and $F^t$ via a feature constraint mechanism.
Finally, the refined geometric feature $F^{g_i}$ and texture feature $F^{t_i}$ are concatenated to directly generate the content-aware viewpoint $\hat{v}_i^o$ for each default viewpoint $v_i$.
These generated viewpoints $\{ \hat{v}_i^o \}$ are used to take place of the default viewpoints $\{ v_i \}$ to obtain projected images for existing projection-related PCQA methods.
To train the proposed CAVGN, we construct a default-optimized viewpoint (DOV) dataset, which contains thousands of paired default viewpoint $v_i$ and their corresponding optimized viewpoint $v_i^o$.
To this end, in order to avoid cumbersome manual selection and labeling, given a default projection region $r_i$ (i.e. viewpoint ${v_i}$) and its corresponding visible part $\mathcal{P}^{v_i}$, we sample a large number of candidate viewpoints, and then propose a self-supervised viewpoint ranking network (SSVRN) to make qualified viewpoint selections under specific constraints.
Our main contributions are as follows:
\begin{itemize}

	\item[$\bullet$] Inspired by the "wooden barrel theory", we propose the first content-aware viewpoint generation network to generate optimized viewpoints for existing projection-related PCQA methods which adopt a content-independent projection strategy via setting viewpoints by default. The proposed CAVGN can directly generate better viewpoints that more fully reflect the authentic quality of degraded point clouds. 

    \item[$\bullet$] To train the proposed CAVGN, we propose a self-supervised viewpoint ranking network (SSVRN) to select the viewpoint with the worst quality projected image to construct a default-optimized viewpoint (DOV) dataset.

    \item[$\bullet$] We select five typical projection-related PCQA methods and three public PCQA datasets to evaluate the effectiveness of the proposed CAVGN. Experimental results show that the projection-related PCQA methods with the optimized viewpoints generated by CAVGN outperform the ones with default viewpoints.

\end{itemize}

\section{Related Works} 
\label{sec:related_works}
In this section, we first review projection-related PCQA methods, and then discuss related works on viewpoint recommendation.

\subsection{Projection-related PCQA}
Projection-related PCQA methods project 3D point clouds onto a number of 2D planes, and then evaluate the quality of projected images using image quality assessment methods \cite{dumic2018artPCQA}. 
These methods can be divided into classical and deep learning-based PCQA approaches.

\textbf{Classical PCQA methods} explore effective, handcrafted feature descriptions for projected images.
Initially, researchers mainly used traditional image processing algorithms, such as Structural Similarity (SSIM) \cite{Tao2009RR-IQA,alex2020SSIM}, Peak Signal-to-Noise Ratio (PSNR), and other indicators, to evaluate the quality of projected images by comparing these indicators.
To improve the effectiveness of features, researchers have also studied different types of feature extraction algorithms and feature combination methods. 
Meynet et al. \cite{meynet220FR-PCQA} obtained good subjective consistency by optimizing the linear combination of geometric and color features.
Yang et al. \cite{yang82020-3d-2d} weighted color texture and depth information on different projected planes to obtain the final quality score.
Zhou et al. \cite{Zhou23rr} obtained salient projection maps through downsampling operations, and then combined content-based similarity and statistical correlation metrics to generate the quality score.
Later, some researchers began to consider that projected images at different positions have different levels of importance. 
Chen et al. \cite{chen9-2021-layer-proj} proposed a full-reference point cloud quality evaluation method based on hierarchical projection, and Wu et al. \cite{wu10-2021-6DoF} proposed a method based on weighted view projection, which assigns corresponding weights to projected images at different positions according to their importance.

\textbf{Deep learning-based PCQA approaches} leverage deep networks to extract intricate features from point clouds, which are more robust than classical PCQA methods.
Nonetheless, they persist in adopting the projection paradigm of traditional methods: by presenting fixed viewpoints to procure projected images of corresponding regions.
Liu et al. \cite{liu12-2021-PQA-net} came up with a deep learning-based no-reference PCQA method called PQA-Net. 
The network parameters are first trained by predicting the types of point cloud distortion, and then joint training is used to predict the quality score.
In their approach, the view space is divided into six areas. The central point of each area is chosen as a fixed viewpoint for obtaining the corresponding projection image.
Tao et al. \cite{tao2021MSFC} projected 3D point clouds onto 2D color projected images and geometric projection images, which employed the same strategy as that of Liu \cite{liu12-2021-PQA-net}, and designed a multi-scale feature fusion network to blindly evaluate visual quality.
Yang et al. \cite{yang13-2022-NR-domationadapt} encoded the projected images of point clouds in different directions into an image, combined them with image datasets, trained the network parameters through domain adaptation methods, and finally successfully predicted the quality score.
Xie et al. \cite{xie2023pmBQA} obtained point cloud quality scores by projecting point clouds into 2D images and fusing multimodal information (texture, normal, depth, and thickness).
Chai et al. \cite{Chai2024} used a degraded-reference branch to measure the distance between complete information and principal component information to characterize the quality state, while the no-reference branch performs multi-scale feature extraction and fusion on the input visual projection.
Finally, a P2IT is proposed to materialize information interaction between planes and points reasonably.
Wang et al. \cite{Wang24eemf} first considered the influence of viewpoint distance and proposed the MOD-PCQA method, which utilizes a three-branch network to extract scale features from different viewpoints, to capture visual features at various granularity levels from fine to coarse.
Zhang et al. \cite{zhang2023mmpcqa} proposed a novel multi-modal learning approach for PCQA, leveraging the advantages of both point cloud and projected image modalities.

Despite the significant advancements achieved by existing projection-related PCQA methods, they all employ a straightforward but content-independent projection strategy, neglecting the distribution of geometric and attribute information of degraded point clouds. 
Therefore, the primary focus of our study is the issue of content-aware projection viewpoint prediction. 

\subsection{Viewpoint Recommendation}

Viewpoint recommendation refers to recommending the optimal viewpoint for users to observe and understand objects or 3D scenes better in different tasks, such as active robot vision \cite{Zengr20cvm} and shape recognition \cite{AbdullahH21}. 
In this section, we provide a brief review of methods for viewpoint recommendation from the following two aspects: quantitative calculation and qualitative analysis.

Quantitative calculations compute values for all viewpoints through explicit metrics.
Following that, they suggest suitable metrics for picking the best one among the candidate viewpoints.
Attneave \cite{attneave1954-visualpercept} argued that the places where the information changes rapidly (i.e., curvature peaks) are the optimality criterion for selecting the optimal viewpoint.
Kamada and Kawai \cite{17Kamada} proposed that a good angle can avoid the degradation of views that project a plane into a line or a line into a point.
Inspired by Kamada and Plemenos' work, Vazquez et al. \cite{20zquez} calculated the entropy using the projected area and number of faces as the criterion to select the optimal viewpoint.

Qualitative analysis is different in that it constructs optimal viewpoint datasets through subjective experiments and then uses deep networks to learn the task's potential optimal criteria.
Marsaglia et al. \cite{24Marsaglia} used subjective tests to build up a dataset containing information about the strengths and weaknesses of all viewpoints to visualize scientific scenes.
He et al. \cite{photo2018He} evaluated each shooting angle subjectively and trained the deep network to effectively evaluate each shooting angle based on the dataset.
Zeng et al. \cite{2020PCNBV} calculated the value of each viewpoint based on the visibility of the model for the reconstruction task offline.
                 
In a word, existing learning-based viewpoint recommendation approaches can select the optimal viewpoint from a set of candidates by understanding the task's inherent needs and establishing evaluation metrics.
However, they all heavily rely on subjective datasets with recommended viewpoints limited to predefined candidate viewpoint sets. These viewpoint recommendation strategies are not applicable to PCQA tasks.
This paper is dedicated to directly generate better viewpoints for given default viewpoints of existing projection-related PCQA methods without the need of a subjective dataset.

\section{Content-Aware Viewpoint Generation Network}
\label{sec:PCQA-BVP: Best Viewpoint Prediction}

\subsection{Overview}

\begin{figure*}[!htbp]
	\centering
	\includegraphics[width=1\textwidth]{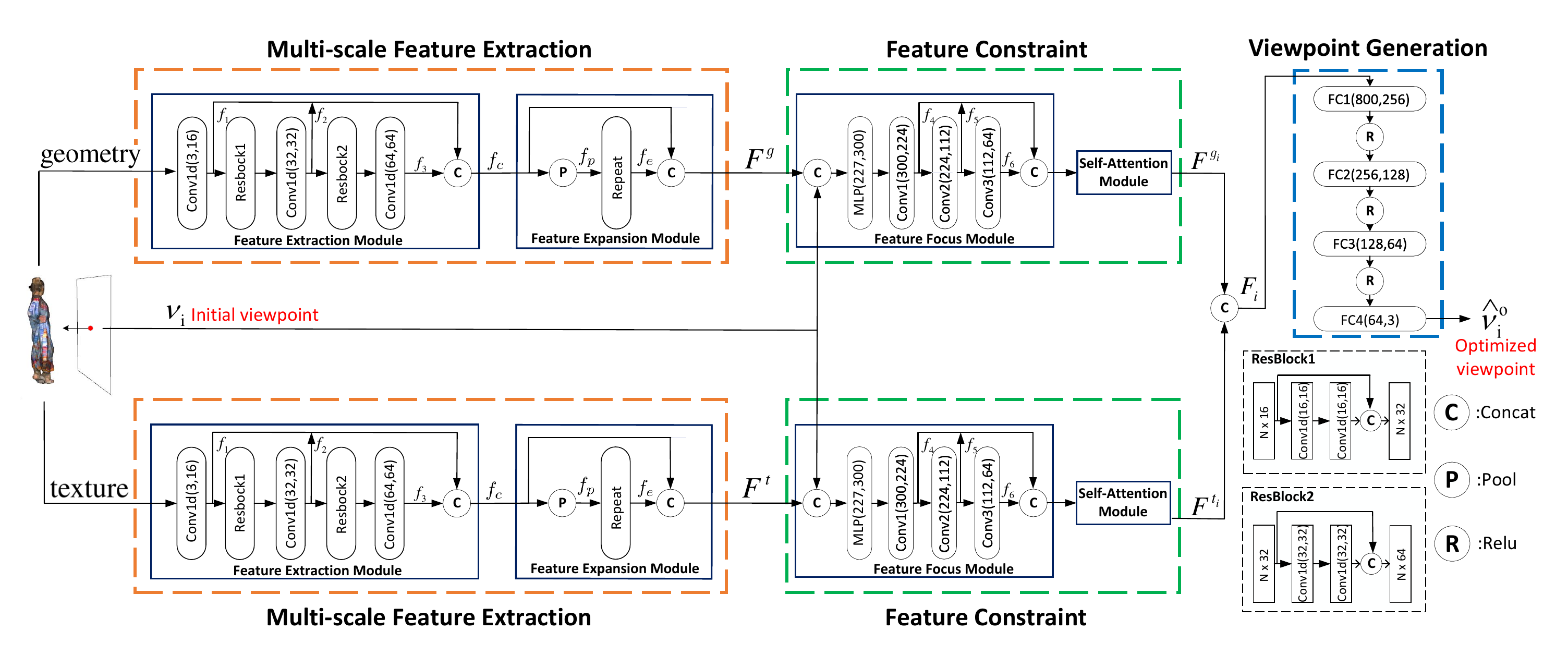}
	\caption{Overview of the proposed content-aware viewpoint generation network (CAVGN). }
	\label{fig:CAVGN}
\end{figure*}

This paper proposes a novel content-aware viewpoint generation network, termed CAVGN, which can directly generate better viewpoints for existing projection-related PCQA methods via taking into account the distribution of geometric and attribute information.
The CAVGN is composed of three parts: multi-scale feature extraction, feature constraint, and viewpoint generation, as shown in Fig. \ref{fig:CAVGN}.
Given the degraded point cloud $\mathcal{P}$ and the default viewpoint sets $\{ v_i \}$, first, the multi-scale feature extraction (MSFE) takes the whole $\mathcal{P}$ as input to extract its geometric feature $F^g$ and texture feature $F^t$, respectively.
Then, for each default viewpoint $v_i$, the feature constraint is specifically designed to focus on the features of the corresponding partial point cloud $\mathcal{P}^{v_i}$ of $v_i$, obtaining the refined geometric feature $F^{g_i}$ and texture feature $F^{t_i}$.
Finally, the viewpoint generation concatenates $F^{g_i}$ and $F^{t_i}$, and generates the optimized viewpoint $\hat{v}_i^o$.

\subsection{Multi-Scale Feature Extraction}

Given the input point cloud $\mathcal{P}$, the MSFE aims to extract multi-scale geometric feature $F^g$ and texture feature $F^t$ of the entire degraded point cloud $\mathcal{P}$, respectively.
It consists of a feature extraction module and a feature expansion module, as shown in Fig. \ref{fig:CAVGN}.

The feature extraction module, which is inspired by the ResSCNN \cite{liuyp23}, is made up of three conv layers and two resblocks.
The output features from each conv layer are initially combined using the concatenation operation,
\begin{equation}
    \left\{\begin{matrix}   
    f_{1} = h_{1}\left ( \cdot \right )  \\    
    f_{2} = h_{2}\left ( r_1\left ( f_{1}  \right )  \right ) \\  
    f_{3} = h_{3}\left ( r_2\left ( f_{2}  \right )  \right ) \end{matrix}\right.,
\end{equation}
where $(\cdot)$ represents the geometry or texture information of $\mathcal{P}$, $h_1$, $h_2$ and $h_3$ are three conv layers, and $r_1$ and $r_2$ are two resblocks.
Then, the feature extraction module outputs the concatenated feature $f_c$:
\begin{equation}
    f_c = [f_1, f_2, f_3].
\end{equation}

The feature expansion module is exploited to fully mine the local and global features of $\mathcal{P}$.
It takes the local features $f_c$ as input, and outputs the expanded features $F$ ($F^g$ or $F^t$).
First, $f_c$ is aggregated into a global feature $f_p$ through the pooling operation.
Afterwards, $f_p$ is transformed into $f_e$ through expand operations
\begin{equation}
    \left\{\begin{matrix} 
  f_{p} = pool(f_{c}) \\  
  f_{e} = \underset{N}{\underbrace{[f_{p},...,f_{p}  ]} } 
\end{matrix}\right. .
\end{equation}
Finally, by concatenating the $f_c$ and $f_e$, the output feature $F$ ($F^g$ or $F^t$) of the expansion module can be obtained:
\begin{equation}
    F=[f_c,f_e].
\end{equation}

\subsection{Feature Constraint}

To learn a better viewpoint for the partial point cloud $\mathcal{P}^{v_i}$ associated with each default viewpoint $v_i$, the feature constraint is designed to refine the geometric feature $F^g$ and texture feature $F^t$ of the entire degraded point cloud $\mathcal{P}$ to pay more attention on the partial point cloud $\mathcal{P}^{v_i}$.
It consists of a feature focus module and a self-attention module.

The feature focus module takes the default viewpoint $v_i$ and the geometric feature $F^g$ or texture feature $F^t$ of the entire degraded point cloud $\mathcal{P}$ as inputs, respectively.
Simultaneously, the relevant initial viewpoint information is incorporated into the $F^g$ or $F^t$.
This helps to express viewpoint related features.
It is composed of a FC layer and three conv layers, with the Leaky Relu as the non-linearity activation function after each layer, as shown in Fig. \ref{fig:CAVGN}. 
The self-attention module \cite{2017attention_vaswani} is then used to better integrate the features from the partial point cloud $\mathcal{P}^{v_i}$ and default viewpoint $v_i$.
At last, we concatenate the refined geometric feature $F^{g_i}$ and texture feature $F^{t_i}$ to obtain the final feature $F_i$, which focuses on the texture and geometric features of the partial point cloud $\mathcal{P}^{v_i}$.

\subsection{Viewpoint Generation}

After obtaining the refined features $F_i$ of the partial point cloud $\mathcal{P}^{v_i}$, the viewpoint generation is designed to directly generate the content-aware viewpoint for $\mathcal{P}^{v_i}$.
Here, four FC layers and three Relu functions are used to extract predictive information and output the optimized viewpoint $\hat{v}_i^o$ of $v_i$.

\subsection{Loss Function}

The optimized viewpoint $\hat{v}_i^o$ needs to be constrained on the projection region $r_i$ where $v_i$ is located. 
In addition, to measure the discrepancy between the generated optimized viewpoint $\hat{v}_i^o$ and its ground truth $v^o_i$, the angle function is employed as the loss function
\begin{equation}
\label{eq:loss}
L\left (v_i^o,\hat{v} ^o_i  \right ) = 1 - cos \left (v_i^o-c,\hat{v} ^o_i-c  \right ),
\end{equation}
where $c$ is the geometric center of $\mathcal{P}$, $\hat{v}_i^o$ is the optimized viewpoint generated by the CAVGN, and $v^o_i$ is the ground-truth optimized viewpoint of ${v}_i$ which comes from the default-optimized viewpoint dataset.

\section{Default-Optimized Viewpoint Dataset}
\label{sec:PCQA-BVDC}

The default-optimized viewpoint (DOV) dataset is constructed in a self-supervised manner to train the proposed CAVGN, which contains thousands of paired default viewpoints $v_i$ and corresponding optimized viewpoints ${v}^o_i$ sampled from different degraded point clouds.
The referred optimized viewpoint ${v}^o_i$ is defined as the viewpoint with the lowest-quality projected image of the partial point cloud $\mathcal{P}^{v_i}$ in this paper.

\subsection{DOV Dataset Construction}
\label{sec4/overview}

To build the DOV dataset, a straightforward way is to manually observe given degraded point clouds from different perspectives and select a qualified viewpoint for each default one.
However, the manual labeling process is time-consuming and laborious.
Therefore, we propose a self-supervised viewpoint ranking network (SSVRN) to build the DOV dataset in a self-supervised manner in this paper, through selecting the viewpoint with the worst quality projected image.

\begin{figure}[!t]
	\centering
	\includegraphics[width=0.3\textwidth]{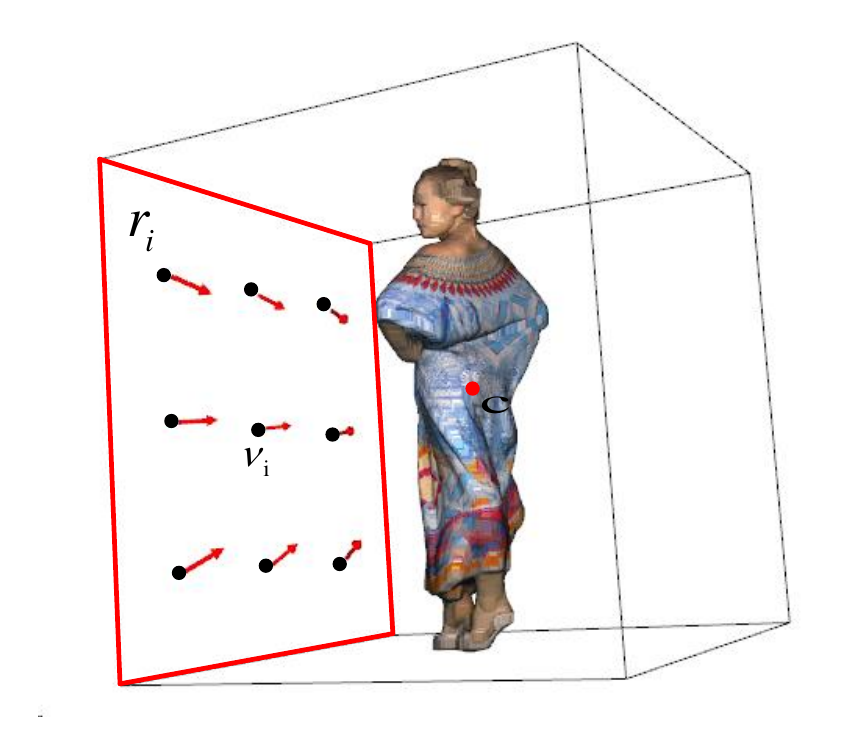}
	\caption{Illustration of candidate viewpoints setting on the projection region $r_i$ under $N_v = 9$.}
	\label{fig:9 viewpoint set}
\end{figure}

Firstly, numerous degraded point clouds are collected.
For each degraded point cloud $\mathcal{P}$, we randomly set a certain number of default viewpoints $\{ v_i \}$ (i.e. projection regions $r_i$). 
Then, for each default viewpoint $v_i$, we set $N_v$ candidate viewpoints $\{ v_i^j \} (1 \leq j \leq N_v)$ within the initial projection region $r_i$ associated with $v_i$, where $\{v_i^j\}$ are uniformly sampled from $r_i$, as shown in Fig. \ref{fig:9 viewpoint set}.
After that, we can get a projected image $I_i^j$ for each candidate $v_i^j$ according to its projection direction $\mathbf{d}_i^j = \overrightarrow{v_i^j - c}$.
To select the qualified viewpoint for the partial point cloud $\mathcal{P}^{v_i}$ from $\{ v_i^j \}$, the proposed SSVRN is employed to rank the projected images $\{ I_i^j \}$.
Finally, the lowest-ranked image is chosen from $ \{I_i^j\}$, and its corresponding viewpoint is selected as the qualified viewpoint $v^o_i$ for $\mathcal{P}^{v_i}$.

\subsection{Self-Supervised Viewpoint Ranking Network}

\begin{figure*}[!h]
	\centering
	\includegraphics[width=0.9\textwidth]{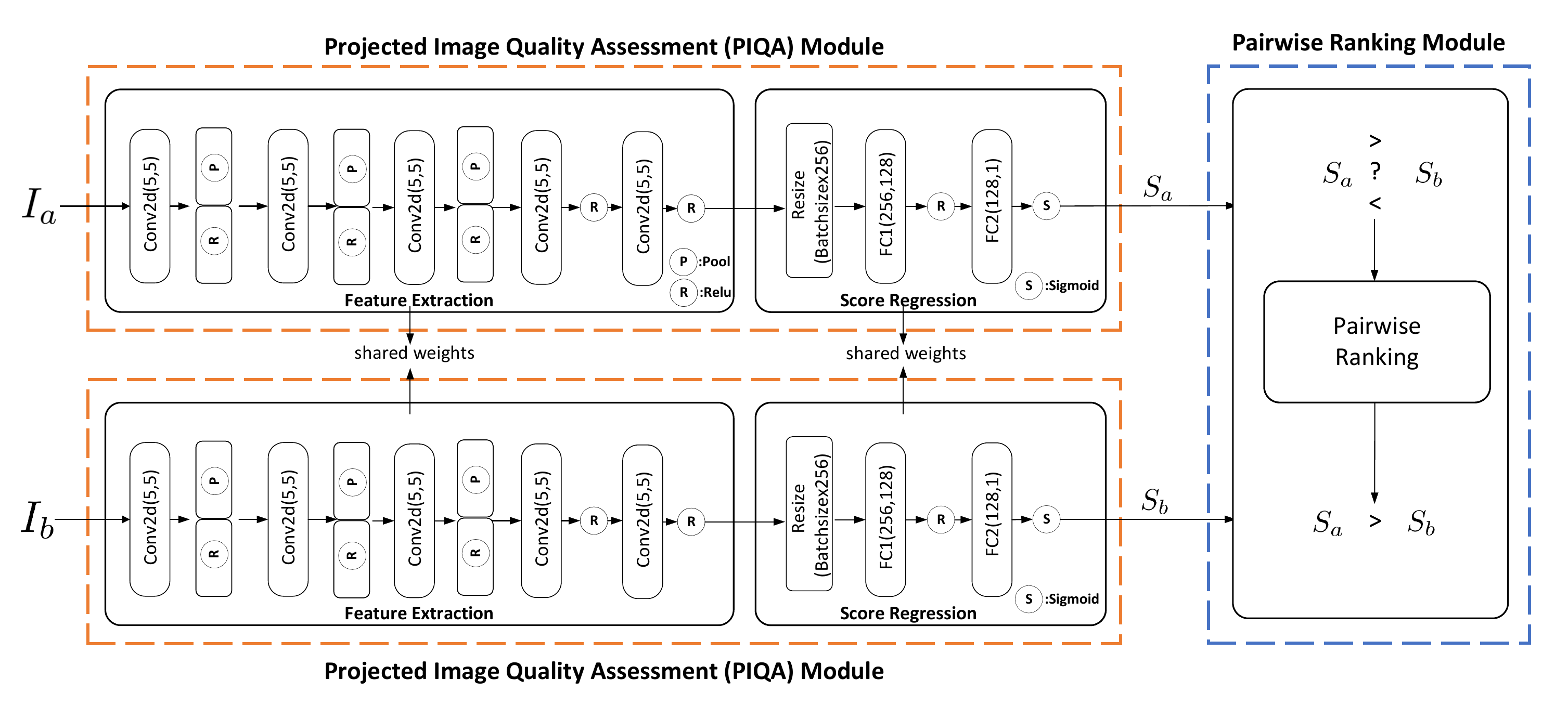}
	\caption{Overview of the proposed self-supervised viewpoint ranking network (SSVRN).}
	\label{fig:Image Rank Net}
\end{figure*}

The proposed SSVRN aims to rank two input projected images observed from different viewpoints of each visible part $\mathcal{P}^{v_i}$ to select a qualified viewpoint with the worst quality projected image for the default viewpoint $v_i$.
It consists of a projected image quality assessment (PIQA) module and a pairwise ranking module, as illustrated in Fig. \ref{fig:Image Rank Net}.
Given two input projected images, the PIQA module first extracts features of projected images and generates quality scores for each image, respectively.
Then, the pairwise ranking module is used to rank the image quality of two input projected images.

\subsubsection{Projected Image Quality Assessment}

The PIQA module extracts image features and produces a quality score for the input projected image, as shown in Fig. \ref{fig:Image Rank Net}.
It is composed of a feature extraction module and a score regression module.
The feature extraction consists of five conv layers, three pooling layers, and uses the Relu as activation function.
The score regression, which consists of two fully connected layers, a Relu activation function and a sigmoid function, is used to obtain the quality score.
The PIQA adopts a dual branch structure, with corresponding modules sharing network parameters, which results in a certain degree of consistency in the extracted features.

\subsubsection{Pairwise Ranking}

The pairwise ranking module is designed to compare the quality of pairs of images ($I_a$, $I_b$).
By utilizing a set of paired images ($I_a$, $I_b$), the pairwise ranking module can compute the probability ${P_{ab}}$ that image $I_a$ is better than image $I_b$,
\begin{equation}
\label{eq:rank equation}
    P_{ab}=\frac{e^{s_a-s_b}}{1+e^{s_a-s_b}},
\end{equation}
where $S_a$ and $S_b$ are the scores of $I_a$ and $I_b$, respectively.
The cross entropy function is used as the loss function,
\begin{equation}
\label{eq:rank_loss}
 L_{ab}=-\bar{P}_{ab}logP_{ab}-(1-\bar{P}_{ab})log(1-P_{ab}) ,
\end{equation}
where $\bar{P_{ab}}$ is the ground truth that image $I_{a}$ is better than $I_{b}$. 

\subsubsection{Self-Supervised Training}
\label{lab:Self-Supervised Training}

To train the proposed SSVRN in a self-supervised way, we first select a large number of point clouds.
Then, each point cloud $\mathcal{P}$ is degraded by a specified distortion under two distortion levels, resulting in $\mathcal{P}_1$ and $\mathcal{P}_2$, respectively.
After that, we randomly set a series of viewpoints for $\mathcal{P}$.
Thus, for each viewpoint, we can get a pair of images through projecting $\mathcal{P}_1$ and $\mathcal{P}_2$ at the same viewpoint, respectively.
The two projected images represent the same content degraded by the same distortion under different distortion levels.
Assume that there are $T$ distortion types, $L$ distortion levels, and $N$ viewpoints for each point cloud $\mathcal{P}$, a total of $N \times T \times C_{L+1}^{2} $ projected image pairs of $\mathcal{P}$ can be generated.
Finally, we employ these projected image pairs to train the SSVRN.

\section{Experimental Results and Analysis}
\subsection{Datasets}

We select the SJTU-PCQA dataset \cite{yang82020-3d-2d}, WPC dataset \cite{su2019perceptual} and LS-PCQA dataset \cite{liuyp23} to build the DOV dataset in this paper.

The SJTU-PCQA dataset encompasses a collection of 9 distinct point clouds, including redandblack, loot, soldier, longdress, Hhi, Shiva, statue, ULB Unicorn, and Romanoillamp. 
Each point cloud is subjected to a comprehensive range of 7 diverse distortion types, incorporating octree-based compression (OT), color noise (CN), down-sampling (DS), geometry gaussian noise (GGN) and 3 kinds of mixed distortion (D+C, D+G, C+G). 
Furthermore, each type of distortion is subdivided into 6 distinct levels.
In total, there are 9 referenced point clouds and $378$ distorted point cloud samples.

The WPC database contains 20 reference point clouds and augments each point cloud with 4 types of distortions, including down-sampling (DS), gaussian white noise (GWN), Geometry-based Point Cloud Compression (G-PCC), and Video-based Point Cloud Compression (V-PCC)).
In total, there are  $20 \times 37 = 740$ distorted point clouds.

The LS-PCQA dataset includes 104 reference point clouds and each reference point cloud is augmented with 31 types of impairments at 7 distortion levels, including color noise(CN), Gamma noise (GN), Poisson noise (PN), Geometry-based Point Cloud Compression (G-PCC), Video-based Point Cloud Compression (V-PCC), etc. 
It is categorized into two parts: 930 models with real MOS and over 22000 models with pseudo MOS.
The 930 models with real MOS are used for our experiments.

\begin{table}[tbp]
	\centering
	\caption{Details of SJTU-PCQA, WPC, and LS-PCQA datasets}
 \resizebox{0.48\textwidth}{!}{
\begin{tabular}{lccccc}
\hline
Dataset   & \begin{tabular}[c]{@{}c@{}}Number of \\ samples\end{tabular} & \begin{tabular}[c]{@{}c@{}}Distortion \\type \end{tabular}                                                          & \begin{tabular}[c]{@{}c@{}}Min.\\ points\end{tabular} & \begin{tabular}[c]{@{}c@{}}Max.\\ points\end{tabular} & \begin{tabular}[c]{@{}c@{}}Mean\\ points\end{tabular} \\ \hline
SJTU-PCQA & 378                                                          & \begin{tabular}[c]{@{}c@{}}OT, CN, DS, D+C,\\ D+G, GGN, C+G.\end{tabular} & 49749                                                 & 2000297                                               & 706684                                                \\
WPC       & 740                                                          & \begin{tabular}[c]{@{}c@{}}DS, GWN, \\ G-PCC, V-PCC.\end{tabular}         & 14901                                                 & 7465623                                               & 1926564                                               \\
LS-PCQA   & 930                                                          & \begin{tabular}[c]{@{}c@{}}CN, GN, PN, \\ G-PCC, V-PCC, etc.\end{tabular} & 13467                                                 & 857966                                                & 710084                                                \\ \hline
\end{tabular}}
	\label{Table:Dataset}
\end{table}

\subsection{Compared Methods}

Five popular and open-source projection-related PCQA methods are selected to evaluate the performance of our approach, which include the full-reference (FR) method SSIM-Net \cite{yang82020-3d-2d}, the reduced-reference (RR) method RR-CAP \cite{Zhou23rr}, and 3 no-reference (NR) methods PQA-Net \cite{liu12-2021-PQA-net}, GMS-3DQA \cite{zhang2023gms}, and MM-PCQA \cite{zhang2023mmpcqa}, respectively. 
Note that the MM-PCQA \cite{zhang2023mmpcqa} is a multimodal-based method.
They all employ the cube-like content-independent viewpoint setting \cite{yang82020-3d-2d,yang13-2022-NR-domationadapt,liu12-2021-PQA-net,zhang2023mmpcqa}.
To investigate the performance of the proposed CAVGN, we also add point-based methods for comparison, including MSE-p2po \cite{Mekuria16p2po}, MSE-p2pl \cite{Tian17p2pl}, PointSSIM \cite{alex2020SSIM}, and 3D-NSS \cite{zhang2022no}.
For SSIM-Net \cite{yang82020-3d-2d} and RR-CAP \cite{Zhou23rr}, we directly use the hyper-parameters provided by the authors.
For PQA-Net \cite{liu12-2021-PQA-net}, GMS-3DQA \cite{zhang2023gms}, and MM-PCQA \cite{zhang2023mmpcqa}, we keep the same network structure and train them by using the dataset mentioned above. 
It is worth noting that the trained network does not achieve exactly the same results as in the referenced papers.
For fair comparisons, we first use the cube-like viewpoint setting to obtain the projected images and compute the quality scores of point clouds.
Then, we use the proposed CAVGN to generate optimized viewpoints. 
Finally, we employ the projected images associated with the generated viewpoints as input for a reevaluation of the quality of point clouds.
In our assessments, we use three fundamental evaluation metrics: 
Pearson linear correlation coefficient (PLCC) \cite{2012—PLCC}, Kendalls rank-order correlation coefficient (KRCC) \cite{liu12-2021-PQA-net}, and Spearman’s rank-order correlation coefficient (SRCC) \cite{yang82020-3d-2d}.

\subsection{DOV Dataset} 

\subsubsection{Experimental Setup}  

To construct the DOV dataset, we take the commonly used cube-like viewpoint setting and set 6 default viewpoints for each point cloud.
We set $N_v = 9$ different candidate viewpoints for each default viewpoint $v_i$, evenly sampled on the projection region $r_i$ centered on $v_i$, as shown in Fig. \ref{fig:9 viewpoint set}.
Thus, there are $9\times 54 \times 7\times C_{7}^{2}=71442$ pairs of projected images from the SJTU-PCQA dataset, $20\times 54 \times 12 \times C_{4}^{2}=77760$ pairs of projected images from the WPC dataset, and $184\times 54 \times C_{6}^{2} =90396$ pairs of images from the LS-PCQA dataset in the DOV dataset.

\begin{table}[!t]
	\centering
	\caption{Ranking accuracy of SSVRN on the testing paired samples }
	\begin{tabular}{lc}
		\hline
		Dataset   & Acc \\ \hline
		SJTU-PCQA & 87\%             \\ 
         WPC      & 84\%        \\ 
         LS-PCQA   & 82\%             \\\hline
	\end{tabular}
	\label{Tabel:Image Ranking Accuracy}
\end{table}

\begin{table*}[!t]
	\centering
	\caption{Experimental results of PQA-Net, GMS-3DQA, SSIM-Net and RR-CAP by using projected images with different quality rankings in the SJTU-PCQA dataset}
\begin{tabular}{ccccccccccccc}
\hline
Quality ranking & \multicolumn{3}{c}{PQA-Net \cite{liu12-2021-PQA-net}}      &  \multicolumn{3}{c}{GMS-3DQA \cite{zhang2023gms}}                   & \multicolumn{3}{c}{SSIM-Net \cite{yang82020-3d-2d} }                        & \multicolumn{3}{c}{RR-CAP \cite{Zhou23rr}}                          \\ \hline
                & PLCC            & SRCC            & KRCC            & PLCC            & SRCC            & KRCC            & PLCC            & SRCC            & KRCC        & PLCC            & SRCC            & KRCC      \\
1               & 0.6671          & 0.6103          & 0.4437          & 0.7569          & 0.6999          & 0.5117          & 0.6190          & 0.6035          & 0.4253          & 0.5838          & 0.5311          & 0.3691          \\
2               & 0.6605          & 0.5744          & 0.4202          & 0.7644          & 0.6167          & 0.4319          & 0.6273          & 0.6118          & 0.4303          & 0.5904          & 0.5837          & 0.4093          \\
3               & 0.6859          & 0.6253          & 0.4558          & 0.7553          & 0.6618          & 0.4841          & 0.6417          & 0.6322          & 0.4454          & 0.5999          & 0.5895          & 0.4156          \\
4               & 0.7219          & 0.6422          & 0.4471          & 0.7749          & 0.6171          & 0.4388          & 0.6450          & 0.6337          & 0.4483          & 0.5939          & 0.5867          & 0.4132          \\
5               & 0.7166          & 0.6202          & 0.4814          & 0.7655          & 0.5473          & 0.3955          & 0.6685          & 0.6595          & 0.4757          & 0.5973          & 0.5892          & 0.4161          \\
6               & 0.7189          & 0.6744          & 0.4786          & 0.7323          & 0.6940          & 0.5111          & 0.6785          & 0.6681          & 0.4808          & 0.6027          & 0.5927          & 0.4185          \\
7               & 0.7403          & 0.6607          & 0.4728          & 0.7899          & 0.7043          & 0.5168          & 0.7062          & 0.6918          & 0.4942          & 0.6064          & 0.5959          & 0.4173          \\
8               & 0.7548          & 0.6819          & 0.4987          & 0.7733          & 0.7096          & 0.5157          & 0.6874          & 0.6760          & 0.4797          & 0.6092          & 0.5979          & 0.4184          \\
9               & \textbf{0.7715} & \textbf{0.7080} & \textbf{0.5094} & \textbf{0.8032} & \textbf{0.7174} & \textbf{0.5347} & \textbf{0.7179} & \textbf{0.7012} & \textbf{0.5118} & \textbf{0.6157} & \textbf{0.6041} & \textbf{0.4238} \\ \hline
\end{tabular}
	\label{Tabel:Different results of 9 projections}
\end{table*}

To train the SSVRN, we randomly use 80\% of the paired images for training and the remaining 20\% for testing.
The Adam optimizer is used to train network parameters.
The learning rate is set to $10^{-4}$ and decreased by a factor of 0.9 every 10 epochs for a total of 100 epochs.

\subsubsection{SSVRN Ranking Performance}

We evaluate the ranking performance of SSVRN by examining whether it can correctly distinguish the sorting of input projected image pairs in the test dataset.
The ranking accuracy is used to evaluate the performance of the SSVRN,
\begin{equation}
    Acc = \frac{D_r}{D_t}, 
\end{equation}
where $D_r$ denotes the number of correct rankings, and $D_t$ is the total number of test image pairs.

Table \ref{Tabel:Image Ranking Accuracy} shows the ranking results on the entire testing dataset. 
From Table \ref{Tabel:Image Ranking Accuracy}, we can see that the sorting performance of SSVRN exceeds 80\% on all datasets, indicating that the proposed SSVRN performs well.
However, as the number of distortion types increases and the difference between distortion levels decreases (e.g., the LS-PCQA dataset), the ranking performance tends to decrease.
The trained SSVRN is used to select the optimized viewpoint $v_{i}^{o}$ from $\{v_i^j\}$ for each default viewpoint $v_i$ and build the DOV dataset, which will be used to train the CAVGN.
Note that the optimized viewpoint here, as discussed in Section \ref{sec4/overview}, corresponds to the viewpoint of the projected image with the lowest score.

\subsubsection{The Worse The Better}

In order to further validate the "wooden barrel theory", we conducted experiments on the SJTU-PCQA dataset.
For each projection region of every point cloud, there are 9 projected images corresponding to the 9 candidate viewpoints of each default viewpoint.
We use the trained SSVRN to rank all of the 9 projected images from the highest (1) quality to the lowest (9) quality. 
After that, we select the projected images with the same quality ordering of each projection region as the input of compared PCQA methods for each test point cloud, and calculate the corresponding quality scores.

Table \ref{Tabel:Different results of 9 projections} shows the performance of four selected PCQA methods by using projected images with different quality rankings. 
It can be observed that the performance of the network is positively correlated with the quality ranking of the projected images, indicating that the image with the lower ranking can faithfully approach the real MOS. 
The reason is that the viewpoints with low-quality projected images focus on more distortion areas that may be obscured, while the viewpoints with high-quality projected images conceal these distortion areas.
These results also demonstrate the "wooden barrel theory" discussed above.

\subsubsection{Subjective User Study}
\label{Subjective}

In this section, we conduct a subjective experiment to verify the consistency between the optimized viewpoints and the preferred viewpoints observed subjectively by humans.
Our subjective experiment is conducted on a 24-inch monitor with a resolution of 1920×1080 with Truecolor (32bit) at 60 Hz and all the steps are compliant with the ITU-R Recommendation BT. 500 \cite{itu}.

Inspired by \cite{liuyp23}, we select a total of 60 models from the SJTU-PCQA, WPC, and LS-PCQA datasets.
As shown in Fig. \ref{fig:9 viewpoint set}, for each sample, we set 9 candidate viewpoints on each face as a test group. 
We invite 15 participants who first received relevant training to familiarize themselves with the selection task. 
In the formal subjective experiment, participants are asked to select the viewpoint with the worst perceived quality from the given 9 candidate viewpoints of each face.
The viewpoint that has been selected the most times for each viewpoint on each face is regarded as the worst viewpoint.
The consistency index $CI$ is used to evaluate the correlation between SSVRN and human selection,
\begin{equation}
    CI = \frac{N_m}{N_t},
\end{equation}
where $N_m$ denotes the number of correct matches between subjectively selected viewpoints and the worst viewpoints ranked by SSVRN, and $N_t$ is the total number of test groups. 
The $CI$ can reach 78\%, which shows that the results of SSVRN are related to the human subjective selections.

\subsection{CAVGN Experiment}
\label{CAVGN Experiment}
\subsubsection{Experiment Setup}

For the CAVGN, we use 80\% of the DOV dataset for training and the remaining 20\% for testing.
The Adam optimization is used to train network parameters, and a batch size of 1 is adopted for training because a larger batch size will exceed the memory capacity.
The initial learning rate is $10^{-5}$ and decreases by a factor of 0.2 every 2 epochs for a total of 30 epochs.

\subsubsection{CAVGN Performance}

To verify the effectiveness of the viewpoints generated by the CAVGN, for each test point cloud, we employ the random viewpoints, the default viewpoints $\{v_i\}$ and corresponding optimized viewpoints $\{\hat{v}_i^o\}$ to assess the performance of PQA-Net \cite{liu12-2021-PQA-net}, GMS-3DQA \cite{zhang2023gms}, MM-PCQA \cite{zhang2023mmpcqa}, SSIM-Net \cite{yang82020-3d-2d}) and RR-CAP \cite{Zhou23rr}, respectively.
The optimized viewpoints $\{\hat{v}_i^o\}$ are generated by the proposed CAVGN.

Table \ref{Table:Performance Verification} illustrates the comparison results with different viewpoints strategies of the above PCQA methods.
From Table \ref{Table:Performance Verification}, we can draw the following conclusions: 
(a) Compared with random viewpoints and default viewpoints, the optimized viewpoints generated by our CAVGN can significantly improve the performance of projection-related PCQA methods.
(b) For the NR methods (PQA-Net \cite{liu12-2021-PQA-net}, GMS-3DQA \cite{zhang2023gms} and MM-PCQA \cite{zhang2023mmpcqa}), the performance improvement is greater than that of the FR (SSIM-Net \cite{yang82020-3d-2d}) and RR (RR-CAP \cite{Zhou23rr}) method.
For example, on the SJTU-PCQA dataset, the PLCC values are improved by 17\% , 12\% and 9\% for the NR methods PQA-Net \cite{liu12-2021-PQA-net} , GMS-3DQA \cite{zhang2023gms} and MM-PCQA \cite{zhang2023mmpcqa}, respectively, while the FR method SSIM-Net \cite{yang82020-3d-2d} and RR method RR-CAP \cite{Zhou23rr} only improved by 8\% and 1\%, respectively.
The reason is that NR methods often lack reference information from the original point cloud, making them more sensitive to the selection of input viewpoints. 
Generating better viewpoints can help NR methods better extract key information, thereby compensating for their inherent disadvantages and improving the evaluation performance.

\begin{table*}[]
\centering
    \caption{Performance of CAVGN on the SJTU-PCQA, WPC and LS-PCQA datasets.}
\label{Table:Performance Verification}
\resizebox{0.9\textwidth}{!}{
\begin{tabular}{clccccccccc}
\hline
          Modal                        &     Methods           & \multicolumn{3}{c}{SJTU-PCQA} & \multicolumn{3}{c}{WPC}                                                              & \multicolumn{3}{c}{LS-PCQA}                                                          \\ \hline
                                  &                & PLCC     & SRCC     & KRCC    & PLCC                       & SRCC                       & KRCC                       & PLCC                       & SRCC                       & KRCC                       \\
\multirow{4}{*}{Point-based}    & MSE-p2po \cite{Mekuria16p2po} & 0.7758   & 0.7153   & 0.5367  & 0.4571                     & 0.4516                     & 0.3128                     &    0.2517                 &     0.1771                 &     0.1284                \\
 & MSE-p2pl \cite{Tian17p2pl} & 0.6705   & 0.6051   & 0.4508  & 0.2511                     & 0.3262                     & 0.2235                     &     0.2230                &   0.1711                   &   0.1273                  \\
                                & PointSSIM \cite{alex2020SSIM}     & 0.7361   & 0.6962   & 0.4933  & 0.4541                     & 0.3431                     & 0.2533                     & 0.2541                     & 0.1601                     & 0.1138                     \\
                                  & 3D-NSS \cite{zhang2022no}         & 0.7382   & 0.7144   & 0.5174  & 0.6514 & 0.6479 & 0.4417 & 0.3241 & 0.2157 & 0.1520 \\ 
                                   \hline
\multirow{5}{*}{\begin{tabular}[c]{@{}c@{}}Projection-related\\(Random viewpoints)\end{tabular}} & SSIM-Net \cite{yang82020-3d-2d}  & 0.5908   & 0.5578   & 0.3867  & 0.6032                     & 0.5756                     & 0.4113                     & 0.2978                     & 0.2698                     & 0.1789                     \\
                                 
                                  & RR-CAP \cite{Zhou23rr}    & 0.5647   & 0.5131   & 0.3460  & 0.4345                     & 0.2467                     & 0.1789                     & 0.2650                     & 0.2576                     & 0.1932                     \\
                                  
                                  & PQA-Net \cite{liu12-2021-PQA-net}    & 0.7122   & 0.6401   & 0.4523  & 0.5175                     & 0.4989                     & 0.3321                     & 0.3309                     & 0.3275                     & 0.2302                     \\
                                  
                                  & GMS-3DQA \cite{zhang2023gms}   & 0.7019   & 0.6898   & 0.5109  & 0.5274                     & 0.5175                     & 0.4345                     & 0.2987                    & 0.2879                     & 0.2352                     \\
                                  & MM-PCQA \cite{zhang2023mmpcqa}   & 0.8459   & 0.7982   & 0.6018  &   0.6987                   &    0.5012                  & 0.6784                   &  0.4351                   &   0.4013                   &   0.2866                  \\
                                 \hline
\multirow{5}{*}{\begin{tabular}[c]{@{}c@{}}Projection-related\\(Default viewpoints)\end{tabular}} & SSIM-Net \cite{yang82020-3d-2d}  & 0.6108   & 0.5713   & 0.3970  & 0.6022                     & 0.5785                     & 0.4109                     & 0.3179                     & 0.3003                     & 0.2042                     \\
                                 
                                  & RR-CAP \cite{Zhou23rr}    & 0.5838   & 0.5311   & 0.3691  & 0.4574                     & 0.2663                     & 0.1870                     & 0.2961                     & 0.2991                     & 0.2014                     \\
                                  
                                  & PQA-Net \cite{liu12-2021-PQA-net}    & 0.7027   & 0.6339   & 0.4481  & 0.5340                     & 0.5020                     & 0.3412                     & 0.3296                     & 0.3344                     & 0.2249                     \\
                                  
                                  & GMS-3DQA \cite{zhang2023gms}   & 0.7201   & 0.6965   & 0.5060  & 0.5572                     & 0.5619                     & 0.4611                     & 0.3101                     & 0.3102                     & 0.2532                     \\
                                  & MM-PCQA \cite{zhang2023mmpcqa}   & 0.8609   & 0.8138   & 0.6315  &  0.7307                 &      0.7118     &         0.5335               &    0.5647     &  0.5367                                &  0.3875                \\
                                 \hline
\multirow{5}{*}{\begin{tabular}[c]{@{}c@{}}Projection-related\\(Generated viewpoints)\end{tabular}} 
                                  & SSIM-Net \cite{yang82020-3d-2d} & 0.6664   & 0.6472   & 0.4598  & 0.6155                     & 0.5929                     & 0.4218                     & 0.3312                     & 0.3210                     & 0.2185                     \\
                                  
                                  & RR-CAP \cite{Zhou23rr}   & 0.5964   & 0.5725   & 0.4020  & 0.4670                     & 0.3330                     & 0.2367                     & 0.3412                     & 0.3399                     & 0.2284                     \\
                                  
                                  & PQA-Net \cite{liu12-2021-PQA-net}  & 0.8207   & 0.7354   & 0.5445  & 0.5840                     & 0.5162                     & 0.3588                     & 0.3617                     & 0.3440                     & 0.2349                     \\
                                 
                                  & GMS-3DQA \cite{zhang2023gms} & 0.8019   & 0.7499   & 0.5616  & 0.5765                     & 0.6168                     & 0.4409                     & 0.3171                     & 0.3280                     & 0.2692                     \\
                                  & MM-PCQA \cite{zhang2023mmpcqa}   & 0.9331   & 0.9020   & 0.7454  &  0.7708                    &       0.7714           &     0.5960                &    0.6214               &  0.5466           &   0.4063            \\
                                  \hline
\end{tabular}}
\end{table*}

\subsubsection{Visualization}

\begin{figure*}[!htb]
\centering  
\subfigure[Visualization of hhi]{
\includegraphics[width=0.48\textwidth]{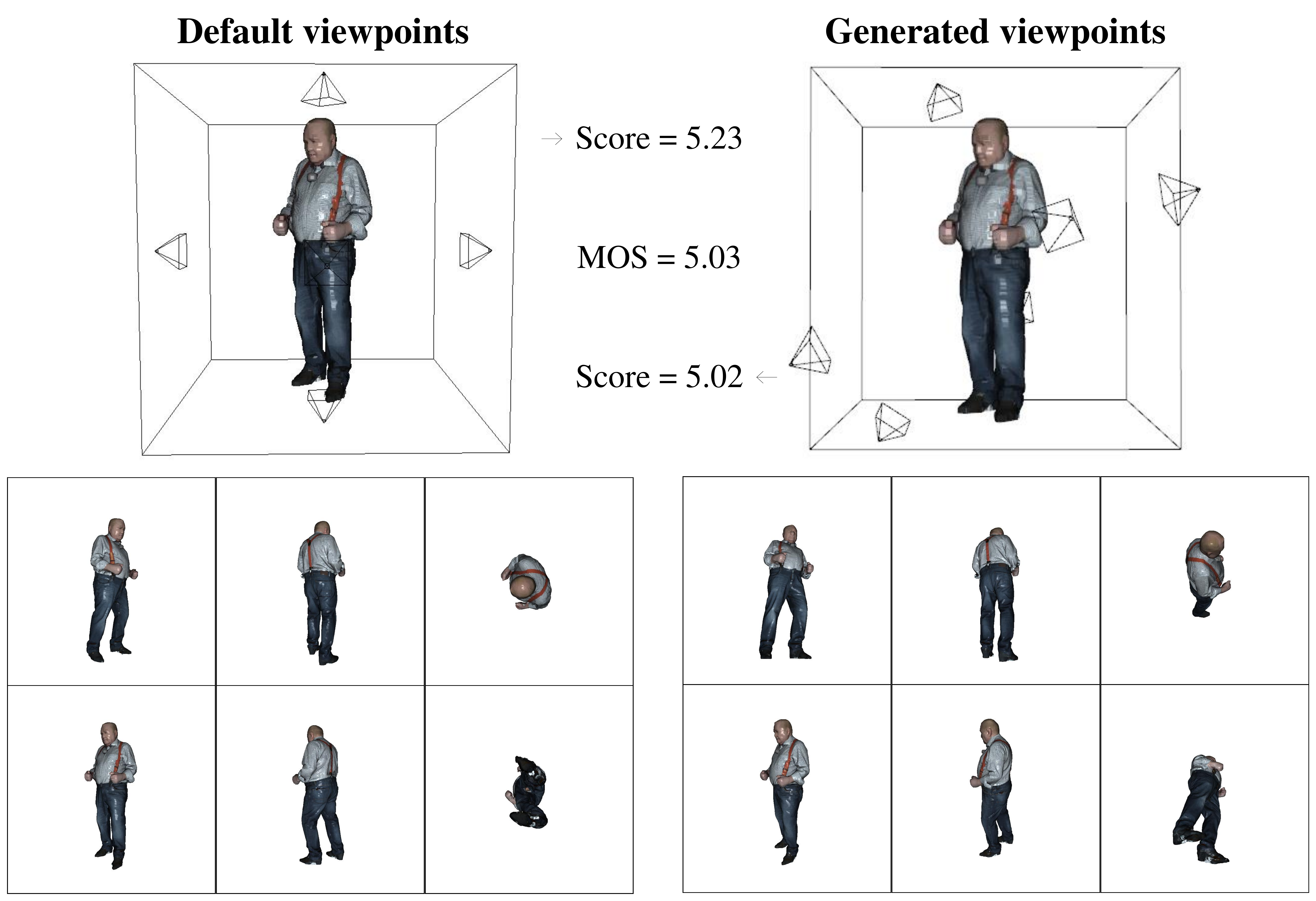}}\subfigure[Visualization of redandblack]{
\includegraphics[width=0.48\textwidth]{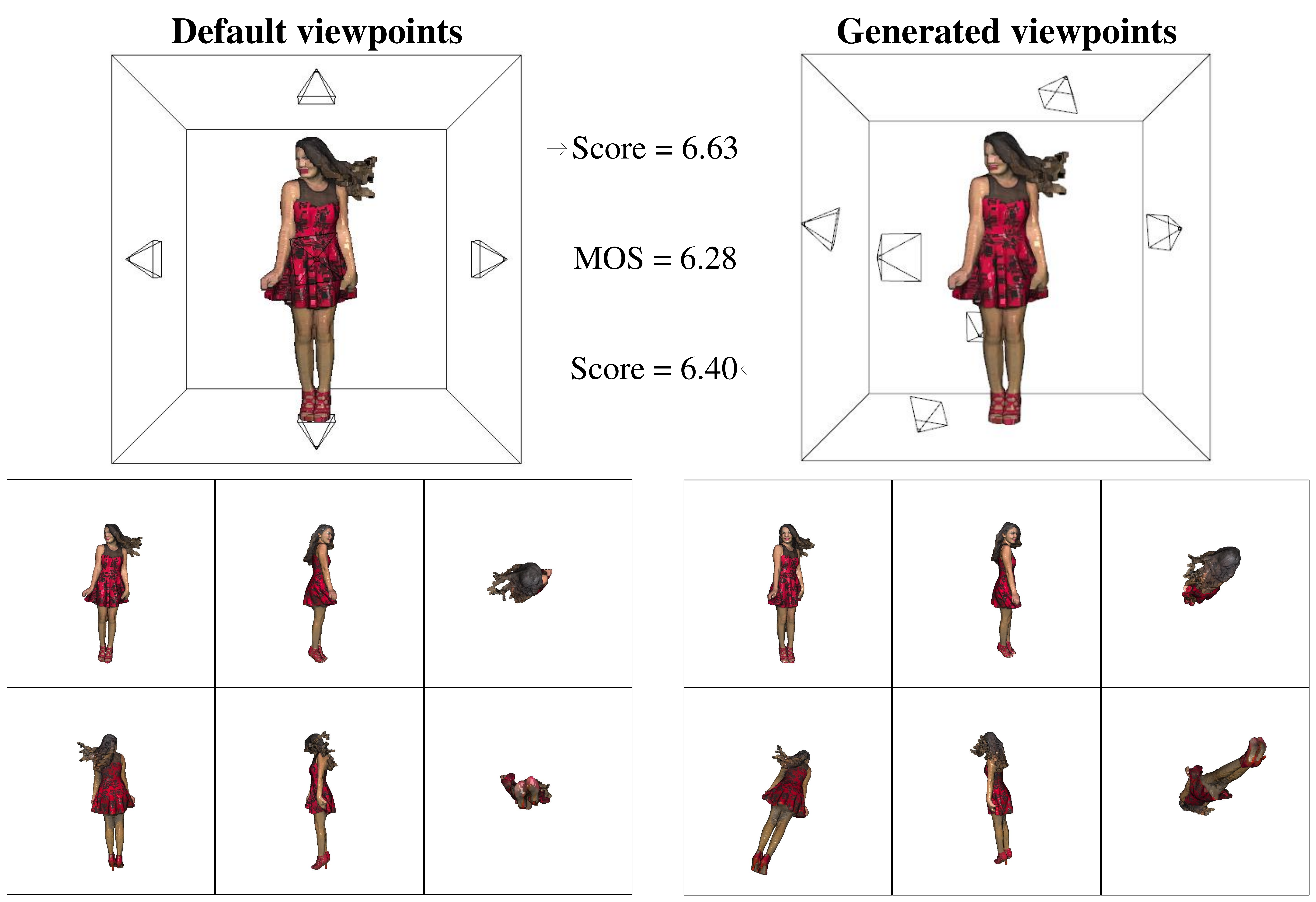}}
\caption{Comparison of default and generated viewpoints of hhi (a) and redandblack (b) in the SJTU-PCQA dataset.}
\label{fig:visualization}
\end{figure*}

\begin{figure*}[!t]
\centering  
\subfigure[Visualization of cake]{
\includegraphics[width=0.48\textwidth]{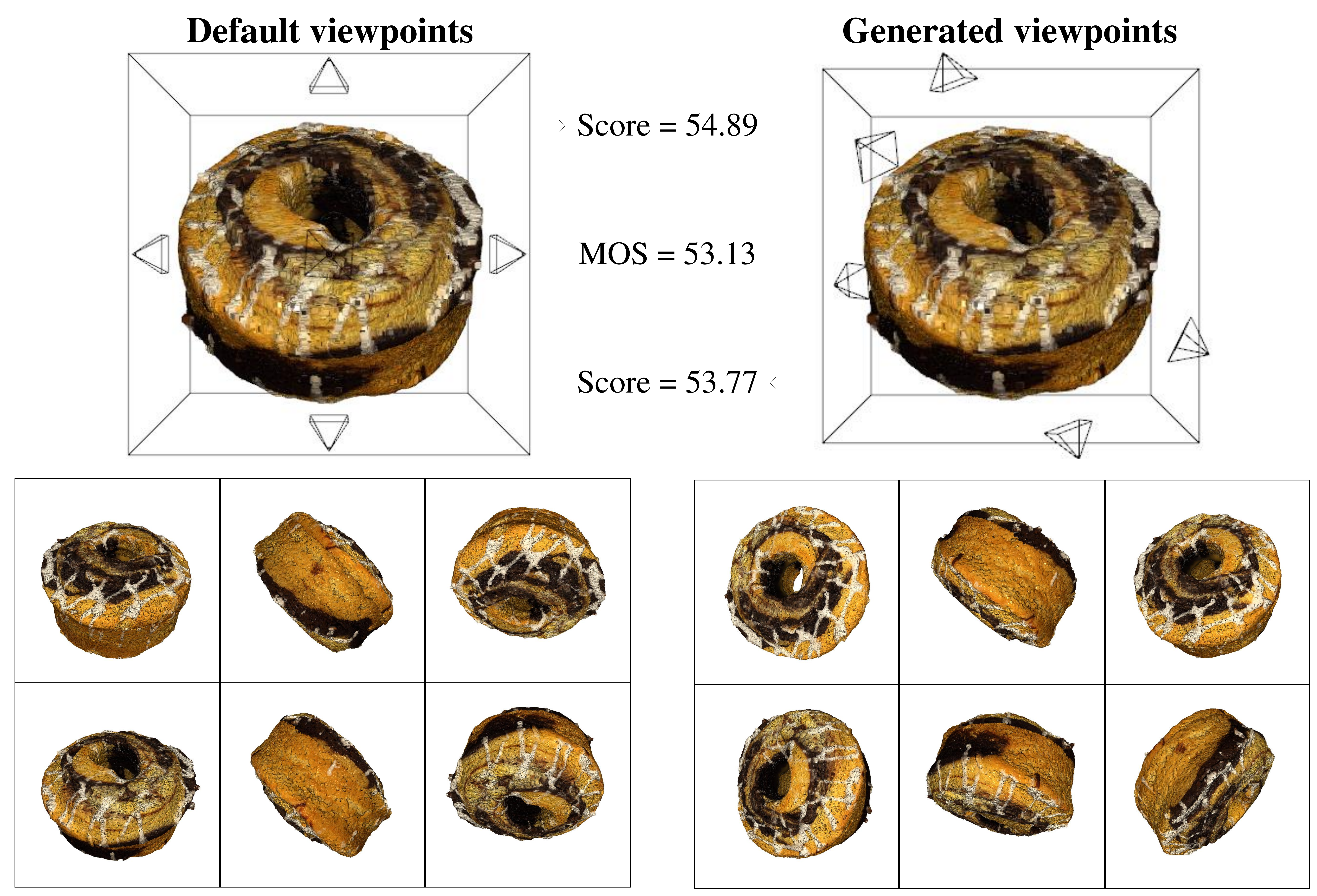}}\subfigure[Visualization of ship]{
\includegraphics[width=0.48\textwidth]{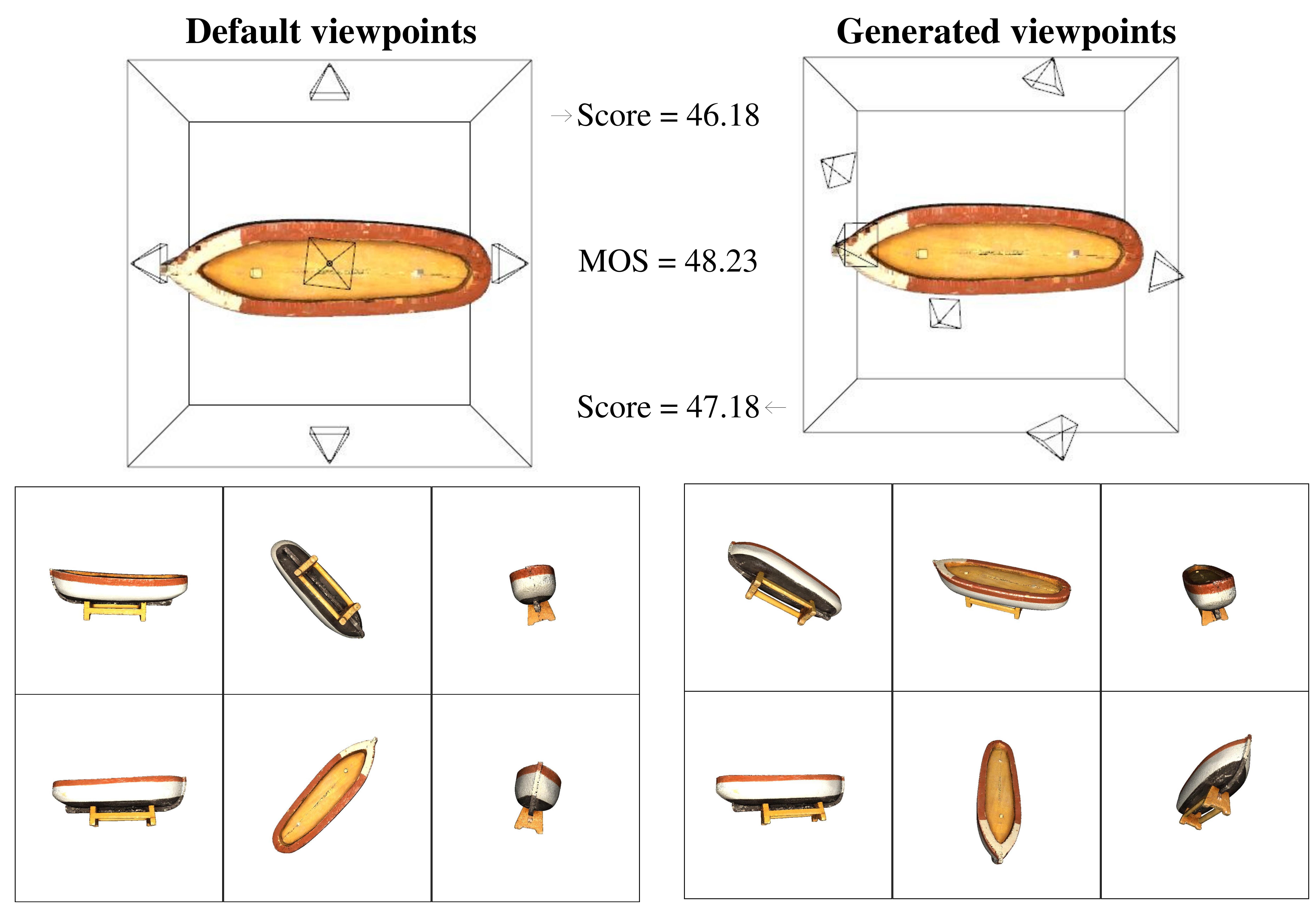}}
\caption{Comparison of default and generated viewpoints of cake (a) and ship (b) in the WPC dataset.}
\label{fig:visualization2}
\end{figure*}

We visualize the default and generated viewpoints of four typical point clouds selected from the SJTU-PCQA dataset (hhi and redandblack) and the WPC dataset (cake and ship), respectively.
The PQA-Net \cite{liu12-2021-PQA-net} is used as the PCQA method to calculate the final quality scores. 
From Fig. \ref{fig:visualization} and Fig. \ref{fig:visualization2} we can see that the generated viewpoints are all correctly distributed on the corresponding projection regions.
Compared to the default viewpoints, these generated viewpoints can provide a more complete picture of the visual quality of the point cloud, especially the distortion regions.
Therefore, the quality scores predicted by using projected images corresponding to the generated viewpoints are closer to the ground truth MOS than the default content-independent viewpoints.

\subsection{Ablation Experiment}

We perform ablation experiments on the candidate viewpoint number $N_v$ and the feature constraint module on the SJTU-PCQA, WPC, and LS-PCQA datasets, respectively.

\begin{table*}[!t]
\centering
\caption{Performance evaluation using different candidate viewpoint schemes. $N_v$ represents the number of candidate viewpoints}
\begin{tabular}{ccccccccccc}
\hline
Dataset  & \multicolumn{3}{c}{SJTU-PCQA}         & \multicolumn{3}{c}{WPC}   &     \multicolumn{3}{c}{LS-PCQA} \\ \hline
  $N_v$                       & PLCC            & SRCC            & KRCC       & PLCC            & SRCC            & KRCC     & PLCC            & SRCC            & KRCC  \\
        $3\times 3$ & 0.7715   & 0.7080 & 0.5094 &0.5735  &0.5231  &0.3632 &0.3614 & 0.3516 &0.2197\\
        $5\times 5$  & 0.7805  & 0.7297 & 0.5216 &0.5810  &0.5404  &0.3841 &0.3726 & 0.3583 & 0.2342\\
        $7\times 7$   & 0.7842 & 0.7322 & 0.5264 &0.5872  &0.5392  &0.3922 &0.3832 & 0.3603 & 0.2453      \\
                             \hline
\end{tabular}
\label{Tabel:Viewpoint Setting Strategy}
\end{table*}

\subsubsection{Numbers of Candidate Viewpoints $N_v$}  

To investigate the effect of $N_v$, we sample different numbers of candidate viewpoints from the projection region $r_i$.
After that, the viewpoint of the projected image with the lowest quality among all candidate viewpoints is selected as the optimized viewpoint.
We employ the PQA-Net \cite{liu12-2021-PQA-net} to evaluate the effectiveness of $N_v$.
Table \ref{Tabel:Viewpoint Setting Strategy} shows the experimental results of $3 \times 3$ viewpoints, $5 \times 5$ viewpoints and $7 \times 7$ viewpoints, respectively.
It can be seen that with the increase of $N_v$, the performance of PQA-Net \cite{liu12-2021-PQA-net} is gradually improved.
This means that the more candidate viewpoints there are, the easier it is to find better viewpoints.
However, for $3 \times 3$ viewpoints, the other two ways give limited performance improvement to the PQA-Net \cite{liu12-2021-PQA-net}, but the computational cost increases exponentially.
Therefore, we propose to use 9 candidate viewpoints to make a trade-off between performance and complexity in this paper.

\subsubsection{Feature Constraint Module}  

To evaluate the performance of the feature constraint module, we combine the output of the MSFE with default viewpoint $v_i$ directly.
Then, the combined feature is used as input for viewpoint generation after pooling.
Table \ref{Ablation Experiment 2} shows the performance of the PQA-Net \cite{liu12-2021-PQA-net} by using the viewpoints generated with and without the feature constraint module, respectively.
It can be observed that the proposed feature constraint module can effectively improve the performance of CAVGN, indicating that it can refine the features extracted by MSFE to focus on the partial point cloud associated with the default viewpoint.

\begin{table*}[!t]
\centering
\caption{Performance evaluation of the feature constraint in CAVGN}
\begin{tabular}{lccccccccc}
\hline
Dataset  & \multicolumn{3}{c}{SJTU-PCQA}         & \multicolumn{3}{c}{WPC}   &     \multicolumn{3}{c}{LS-PCQA} \\ \hline
Types                      & PLCC            & SRCC            & KRCC       & PLCC            & SRCC            & KRCC     & PLCC            & SRCC            & KRCC          \\ 
Without feature constraint & 0.7032          & 0.7053          & 0.5095    & 0.5262          & 0.4694          & 0.3247 & 0.3014 & 0.2536 & 0.1724      \\
With feature constraint              & \textbf{0.8207} & \textbf{0.7354} & \textbf{0.5445} & \textbf{0.5840} & \textbf{0.5162} & \textbf{0.3588} &\textbf{0.3617} & \textbf{0.3440} & \textbf{0.2349}\\ \hline
\end{tabular}
\label{Ablation Experiment 2}
\end{table*}

\subsection{Cross-dataset Validation Experiment} 

To verify the generalization ability of the proposed method, we conduct cross-dataset validation experiments. 
Specifically, the CAVGN is independently trained on four datasets: SJTU-PCQA, WPC, LS-PCQA, and DOV. 
For each dataset, 80\% of the data is used for training, while the remaining 20\% is reserved for validation. 
The trained models are then tested on the entire SJTU-PCQA, WPC, and LS-PCQA datasets, respectively.
The PQA-Net \cite{liu12-2021-PQA-net} is used for evaluation. 
From Table \ref{cross-dataset}, it can be seen that there is no significant performance difference between the cross-dataset settings, indicating that the proposed CAVGN has good generalization ability.
Additionally, the proposed CAVGN trained on the DOV dataset achieves the best results, demonstrating that the DOV dataset can effectively improve the generalization ability of CAVGN.

\begin{table}[!t]
\centering
\caption{Performance results of cross-dataset evaluation}
\label{cross-dataset}
\resizebox{0.48\textwidth}{!}{
\begin{tabular}{lcccccc}
\hline
\multirow{3}{*}{Training} & \multicolumn{6}{c}{Testing}                                                              \\ \cline{2-7} 
                       & \multicolumn{2}{c}{SJTU-PCQA} & \multicolumn{2}{c}{WPC} & \multicolumn{2}{c}{LS-PCQA} \\
                       & PLCC          & SRCC          & PLCC       & SRCC       & PLCC         & SRCC         \\ \hline
SJTU-PCQA              & 0.7639        & 0.7289        & 0.5457     & 0.4963     & 0.3379       & 0.3321       \\
WPC                    & 0.7578        & 0.7163        & 0.5699     & 0.5024     & 0.3218       & 0.3139       \\
LS-PCQA                & 0.7516        & 0.7201        & 0.5579     & 0.5012     & 0.3578       & 0.3198      \\ 
DOV  & \textbf{0.8207} & \textbf{0.7354}  & \textbf{0.5840} & \textbf{0.5162}  &\textbf{0.3617} & \textbf{0.3440}  \\
\hline
\end{tabular}}
\end{table}

\subsection{Time and Complexity Analysis}

In this section, we analyze the computational efficiency of the proposed CAVGN. Three metrics (parameters, inference time, and GPU memory) are used for analysis. 
The experiment is based on the Inter I9-14900K and Nvidia RTX4090D platforms.
The parameters are 4.54M and the average resource consumption on the SJTU-PCQA, WPC, and LS-PCQA datasets is shown in Table \ref{resource analysis}.
It can be seen that the average resource consumption on the WPC dataset is almost twice that of the SJTU-PCQA and LS-PCQA datasets.
This is because the average number of points in the WPC dataset is about 2 million, which is about twice the average number of points in SJTU-PCQA and LS-PCQA datasets.

\begin{table}[!t]
\centering
\caption{Computational resource analysis of CAVGN on SJTU-PCQA, WPC, and LS-PCQA datasets}
\label{resource analysis}
\begin{tabular}{lcc}
\hline
 Dataset  & Inference time (s) & GPU memory (GB) \\ \hline
SJTU-PCQA    &   2.78            & 10            \\
WPC          & 6.27              & 20            \\ 
LS-PCQA      & 2.15              & 7           \\
\hline
\end{tabular}
\end{table}

\subsection{CAVGN Vs SSVRN}

The proposed CAVGN, which is a generative approach, is designed to generate better viewpoints directly and efficiently.
The presented SSVRN is employed to rank two input projected images with different viewpoints and then build the DOV dataset to train the CAVGN in this paper.
In essence, as with the CAVGN, the introduced SSVRN can also be used to select qualified viewpoints from a large number of candidate viewpoints for existing projection-related PCQA methods, which is demonstrated in Table \ref{Tabel:Different results of 9 projections}.
However, the SSVRN is a search-based approach and suffers from high computational cost, such as candidate viewpoints generation, image projection and ranking.
On the contrary, the CAVGN can achieve efficient viewpoint generation without these computationally expensive operations.
Therefore, we can choose the appropriate method as needed in practice.

\section{Conclusions and Future Work}
\label{sec:Conclusions}

In this paper, we propose a novel content-aware viewpoint generation network to address the problem of content-independent viewpoint setting of existing projection-related PCQA methods.
Besides, we also present a self-supervised viewpoint ranking network to construct a default-optimized viewpoint dataset to train the proposed CAVGN.
To the best of our knowledge, CAVGN is the first content-aware viewpoint generation method.
Our experiments demonstrate that the viewpoints generated by the proposed CAVGN can yield significant improvements in the performance of existing projection-related PCQA approaches.

The CAVGN, as a kind of point-based PCQA method, may increase computational complexity and resource consumption, particularly when processing large-scale point cloud data, which could limit its application in resource-constrained environments. 
Future work can be carried out from two aspects: point cloud preprocessing (e.g., two-stage sampling \cite{Zhu243dta}, and the sub-model strategy \cite{zhang2023mmpcqa} ) and network structure (e.g., sparse convolution \cite{liuyp23}).
Besides, we primarily focus on viewpoint selection to enhance the effectiveness of projection-related PCQA methods in this study. However, the scales also play a significant role in the projection process, as different scales can capture varying levels of detail and global contexts \cite{Wang24eemf}. 
In the future work, we aim to explore adaptive strategies for the scale selection, allowing the method to dynamically adjust scales based on the characteristics of point clouds or task requirements.

\ifCLASSOPTIONcaptionsoff
  \newpage
\fi

\begin{IEEEbiography}[{\includegraphics[width=1in,height=1.25in,clip,keepaspectratio]
{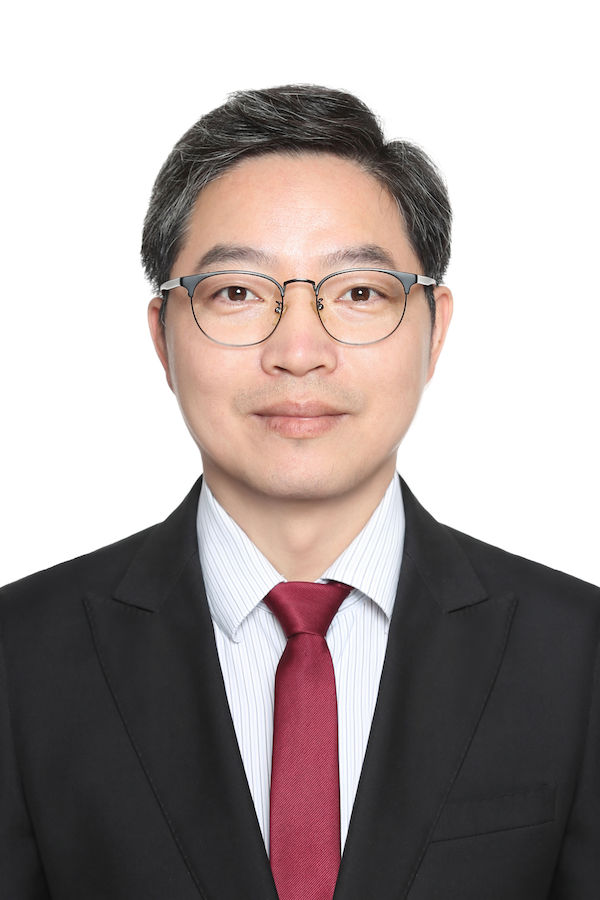}}]{Zhiyong Su}
is currently an associate professor at the School of Automation, Nanjing University of Science and Technology, China. He received the B.S. and M.S. degrees from the School of Computer Science and Technology, Nanjing University of Science and Technology in 2004 and 2006, respectively, and received the Ph.D. from the Institute of Computing Technology, Chinese Academy of Sciences in 2009. His current interests include computer graphics, computer vision, augmented reality, and machine learning.
\end{IEEEbiography}

\begin{IEEEbiography}[{\includegraphics[width=1in,height=1.25in,clip,keepaspectratio]{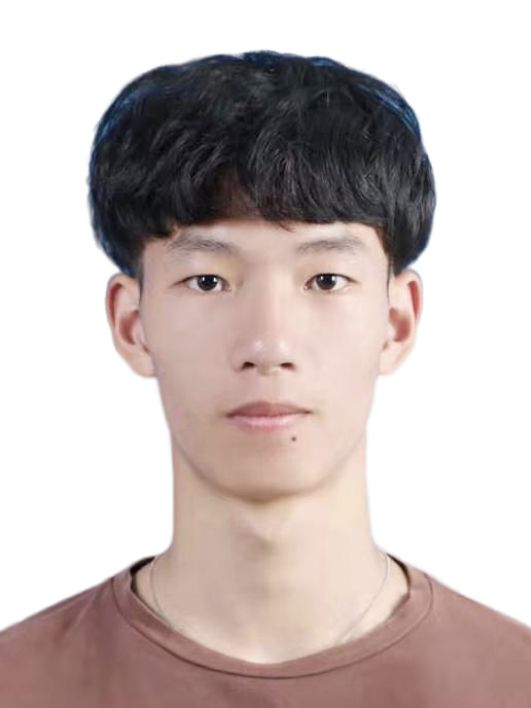}}]{Bingxu Xie} received the B.S. degree from Jiangnan University, Jiangsu, China in 2023 and now studies at Nanjing University of Science and Technology. His research interests include point cloud quality assessment and machine learning.
\end{IEEEbiography}

\begin{IEEEbiography}[{\includegraphics[width=1in,height=1.25in,clip,keepaspectratio]{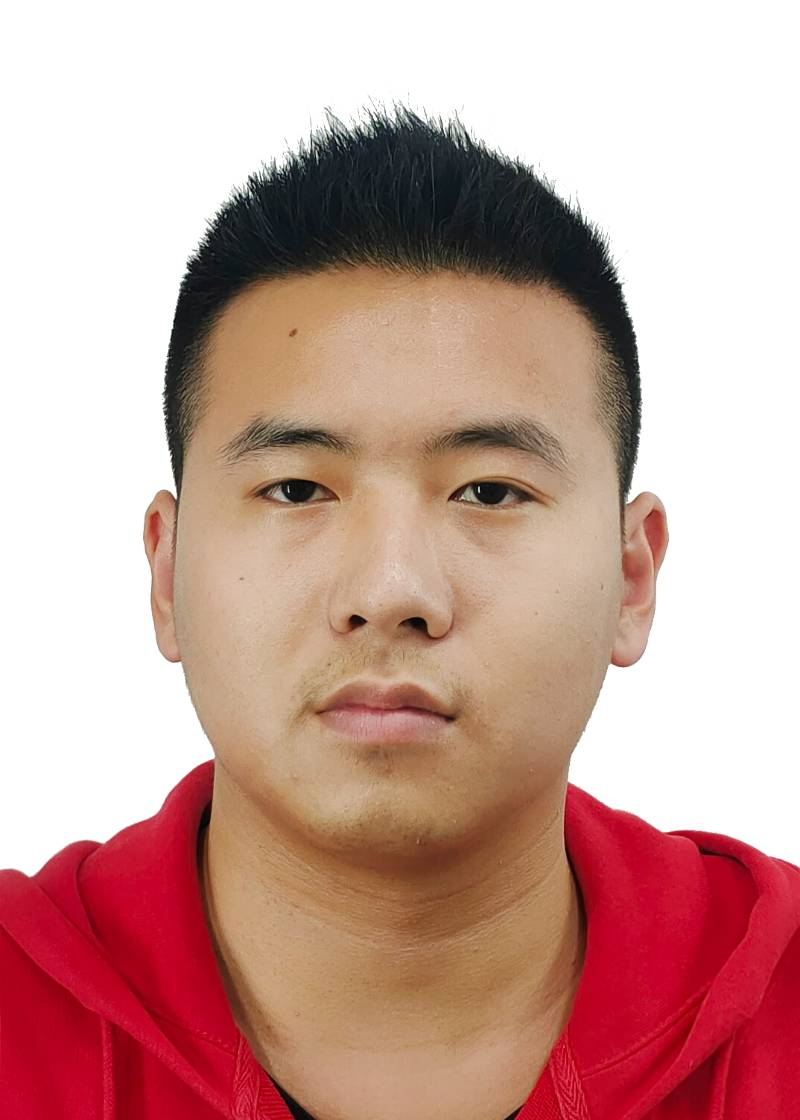}}]{Zheng Li}
	received the B.S. degree from Xuzhou University of Technology, Jiangsu, China in 2022 and now studies at Nanjing University of Science and Technology. His research interests include quality assessment of point clouds.
\end{IEEEbiography}

\begin{IEEEbiography}[{\includegraphics[width=1in,height=1.25in,clip,keepaspectratio]{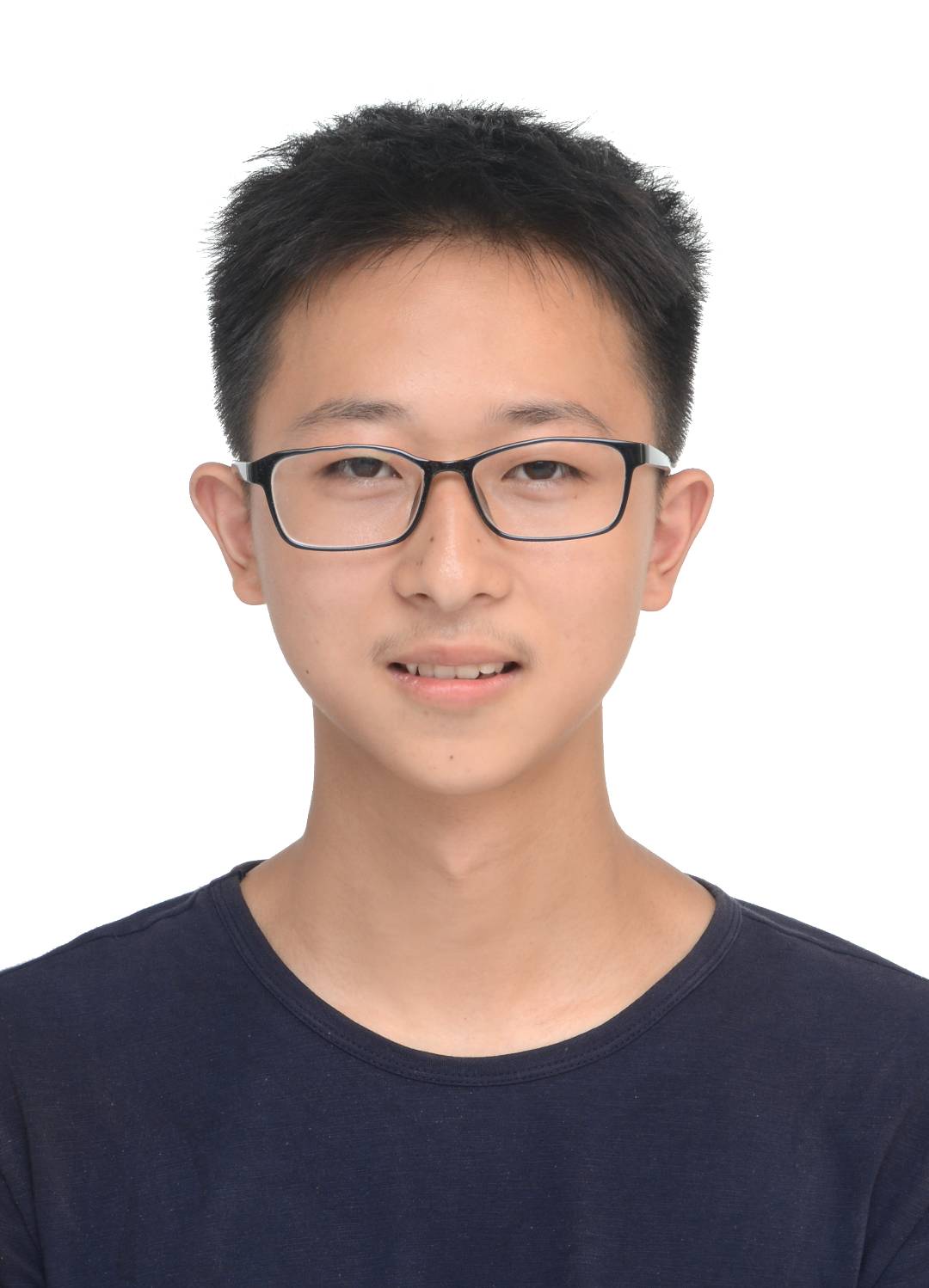}}]{Jincan Wu}
received the B.S. degree from Nanjing University of Science and Technology, Jiangsu, China in 2023 and now studies at Nanjing University of Science and Technology. His research interests include mesh quality assessment and 3DGS quality assessment.
	
\end{IEEEbiography}

\begin{IEEEbiography}[{\includegraphics[width=1in,height=1.25in,clip,keepaspectratio]
{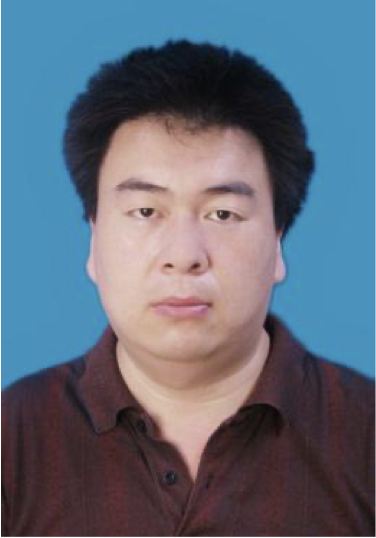}}]{Weiqing Li}
is currently an associate professor at the School of Computer Science and Engineering, Nanjing University of Science and Technology, China. He received the B.S. and Ph.D. degrees from the School of Computer Sciences and Engineering, Nanjing University of Science and Technology in 1997 and 2007, respectively. His current interests include computer graphics and virtual reality.
\end{IEEEbiography}

\vfill

\end{document}